# A Structured Approach to Trustworthy Autonomous/Cognitive Systems

Henrik J. Putzer

cogitron GmbH, Stefaniweg 4, 85652 Pliening, Germany, henrik.putzer@cogitron.de

Ernest Wozniak

fortiss GmbH, Guerickestr. 25, 80807 Munich, Germany, wozniak@fortiss.org

**Abstract**

Autonomous systems with cognitive features are on their way into the market. Within complex environments, they promise to implement complex and goal oriented behavior even in a safety related context. This behavior is based on a certain level of situational awareness (perception) and advanced decision making (deliberation). These systems in many cases are driven by artificial intelligence (e.g. neural networks). The problem with such complex systems and with using AI technology is that there is no generally accepted approach to ensure trustworthiness. This paper presents a framework to exactly fill this gap. It proposes a reference lifecycle as a structured approach that is based on current safety standards and enhanced to meet the requirements of autonomous/cognitive systems and trustworthiness.

## Introduction

Autonomous/cognitive systems are taking over safety-relevant tasks in many industries, for example in the medical, the automotive, or the aviation industry. Their usage extends beyond limited operational environment into highly complex one, where engineered functions operate autonomously. This resulted in a term, recently used quite extensively, autonomous systems. These autonomous/cognitive systems (A/C-system), as they are called in this work, due to the high criticality of functions that they implement, need some form of qualification or even certification before being approved for the market. However, even in cases where a formal certification is not legally required, national and international standards provide guidelines and best practices that aim at minimizing unacceptable risks for bringing products to market. There are many standards as such. They either object on disjoint aspects of system development, complementing one another, or significantly overlap, presenting however different strategies to reach desired goals. An example of the last this could be trustworthy system design.

Standard IEC 61508 [Commission (2010)] is a generic standard to address functional safety of electric, electronic and programmable elements for all industries. It looks at risks that evolve from malfunctioning and does not cover the intended performance. In IEC 61508 the underlying approach is to enable qualification by providing a safety case together with the product. To generate the safety case, which is a structured argumentation, IEC 61508 defines a structured approach called safety lifecycle.

An adaptation of this standard for the Automotive Industry is the ISO 26262 [ISO (2011)] which addresses functional safety of electric, electronic and programmable elements and these risks that evolve from malfunctioning. Complementary to this is ISO/PAS 21448 [ISO (2019)] which is also targeting road vehicles. It is focused around the absence of unreasonable risk due to hazards resulting from functional insufficiencies (performance) of the intended functionality or by reasonably foreseeable misuse by persons. These are referred to as the Safety Of The Intended Functionality (SOTIF).

This paper argues that these standards do not sufficiently cover the aspects of A/C-systems. First shortage is due to the important property of the A/C-systems which is interactive behavior with a complex environment. This interaction needs to account for constantly changing surrounding conditions, and consider scenarios that were not even envisioned when the system was designed. This paper argues on the need of a new phase in the overall system design lifecycle, which would encompass this concern.

The second deficiency is due to the potential usage of AI technology to implement autonomous behavior of A/C-systems. NNs which are the main exponent of AI technology represent a promising approach to cope with the complexity of these future systems, creating at the same time new demands. While progress in AI is accelerating, standardization efforts for safety-critical systems are not keeping up. For example, ISO 26262 ("Road vehicles – Functional safety") does not define how AI can be safely applied in its domain. This even holds for the updated version that is currently under revision. Yet, industry and academia is researching and developing self-driving cars – of course using AI technology. Therefore, it is necessary to develop a structured methodology that ensures a sufficient quality level when developing systems involving AI.

Considering above shortages in current state of the art, the main goal of this work is to develop a structured approach, here called reference lifecycle, which ensures a sufficient quality level. The last states core part of the evolving DKE/AK 801.0.8 standard [DKE (2019)], focused on the specification of trustworthy A/C-systems. The main criterion for concepts introduced here is their flexibility, so that it is possible to integrate them into multiple different existing (safety) standards. It is believed that such an approach can increase chances for an acceptance of these new concepts by the industry and therefore lead to a faster publication of the new standard.

This work is structured in a following way: Next section presents related work, focusing on the safety aspect (a crucial ingredient of trustworthiness concept defined in this paper) of complex systems, and the problem of their qualification within engineering approach. Section Key Concepts defines key concepts used throughout the entire paper. Following section discusses the main idea which is a reference lifecycle and overall structured methodology to develop trustworthy A/C-systems. Finally, the last section concludes this paper and discusses next steps towards a complete specification of the structured approach.

## Related Work

Engineering of trustworthy A/C-systems introduces new set of problems and challenges which mostly result from the open environment in which these systems operate. Closed environment is an implicit assumption within currently existing standards such as [Commission (2010)], [ISO (2011)] and [ISO (2019)]. They all shape an engineering approach towards development of safety-critical systems, but without explicit consideration of autonomous and cognitive behavior. This has resulted in attempts to implicitly use such standards, referring mainly to their possible connections with AI, which is widely researched as a technology to implement A/C-systems. Relatively recent works try to reason about a safety of AI based solutions by referring into the mentioned standards. Work from [Rick Salay (2018)] identifies 34 methods related to unit development in ISO 26262 [ISO (2011)] part 6 (i.e. part related to Software development) where 27 of them are highly recommended for ASIL D (Automotive Safety Integrity Level – level of criticality, where D represents the highest level). Authors show that most of these methods could be applied to machine learning hence increasing its reliability. For example, initialization of variables. However, 7 of these methods require adaptation, e.g. semi-formal notations. Henriksson et al. [Henriksson (2018)] shows how to proceed with such adaptations. [Gosavi and Conrad (2018)] is also attempting to extract methods from ISO 26262 which could be used, so that safety can be introduced in autonomous and semi-autonomous vehicles. Common thing among these works is that they all strive to use the standard as it is and see how existing techniques, coined mainly for improving safety of Software (SW), could be reused with slight adaptations, if necessary.

Traditionally, it was considered best practice, not to utilize AI (especially machine learning) for safety-relevant tasks. For example, Bergmiller's work on functional safety in drive-by-wire vehicles states that neural networks are unsuitable for such a system [Bergmiller (2015)]. He cites mainly their lack of interpretability and states that their downsides apply to most other machine learning techniques as well.

Kurd et al. [Kurd, Kelly, and Austin (2006)] propose a path towards certifying neural networks for safety-relevant systems. They propose hybrid networks where symbolic knowledge is inserted into a neural network and after the learning process, refined symbolic knowledge is extracted. This approach avoids the black-box view of traditional neural networks at the cost of having to solve the additional problem of extracting knowledge from the network. The latter is known to be NP-hard. Furthermore, Kurd et al. discusses safety criteria of neural networks and present them in the form of goal structuring notation. Importantly, authors present also a safety lifecycle to be applied at the technology level, based on the 'W' model, as they call it. It is reasoning about concerns which result from the usage of hybrid networks. For example, one of the steps in the lifecycle of Kurd et al. is called Initial knowledge where initial knowledge is converted into symbolic forms. Framework proposed in this work does not collide with the 'W' model. In fact, approach of Kurd et al. could be easily integrated into framework presented in this work through so called concept of a blueprint, explained later in this work. Kurd's PhD thesis [Kurd (2005)] contains a more detailed discussion as well as a survey on neural networks in safety critical systems.

Another proposal towards certifiable neural networks comes from Morgan et al. [Morgan et al.(1996)]. The difference is that their approach heavily focuses on certifying the process of training the network instead of certifying the network itself. They also raise a set of questions and guidelines that a corresponding standard should answer. Rodvold [Rodvold (1999)] proposes a different development process for neural networks that resembles the waterfall model of traditional software development.

Pulina and Tacchella [Pulina and Tacchella (2010)] present an approach, where a neural network is modelled via Boolean combinations of linear arithmetic constraints in such a way, that the constraints are consistent if and only if the network is safe. Therefore, deciding the safety of the neural network can be answered by finding a satisfying assignment for the constraints.

What is characteristic about all these works, is that they are focusing on one, very particular issue, i.e. employment of AI, especially NNs, in the context of safety critical systems. Even if some of them explicitly refer to standards which define structured approach, their reasoning is restricted to the level of AI introduction. They don't consider the problem from the broader perspective, i.e. how A/C-systems design could influence all the levels of an engineering approach, not just the level of a specific technology, i.e. AI, SW or HW technology.

## Key Concepts

Structured approach to dependable cognitive systems and dependable AI applications discussed in this work is based on several key concepts. They are described in this section and will be referenced by the description of the trustworthiness reference lifecycle presented in the follow-up section.

### Autonomous/Cognitive System

The term autonomous/cognitive system is used as an abstraction from AI or NNs, to focus on functionality and behavior. With this term, the overall system is addressed, not only the algorithms as the core of behavior generation. Actually, NNs or more broadly, AI algorithms, might be part of a cognitive system but are no precondition as long as the systems shows autonomous/cognitive behavior. This means that on the technical level, it is left to the people to decide if standard SW/HW processes are enough or there exist special needs that one could deliver only with AI method. Another term that was introduced, and which similarly as A/C-systems describe systems that bear behavior, one would normally associate with human behavior in terms of complexity, these are open context systems [Burton, Gauerhof, and Heinzemann (2017)]. During the design the capabilities and skills to generate the behavior of an A/C-system should be derived from the task share with the user and from the interaction with other entities in the environment. These definitions form a separate and early design phase in a design process of cognitive systems.

The behavior of a cognitive system is defined by the inputs and outputs of the system. For an illustration refer to Figure 1; here the cognitive system is represented as ovals on the right (the so called body of the cognitive system) in opposite to the world or environment where the user can be found. The environment and the A/C-system are connected via inputs and outputs. Between these input and output arrows, the behavior can be measured (Behaviour arrow). The behavior is conceptually a function taking the input and generating the output, considering also the internal state of the cognitive system.

The behavior generating function is decomposed into three subfunctions (the so-called knowledge transformators): Perception, Deliberation and Execution – which is similar to the decomposition paradigm of "sense-plan-act" in the robotics domain. These subfunctions are further detailed into skills. For example, the perception could be structured along skills of "multi sensor data fusion", "object detection", "object classification" and "intent recognition".

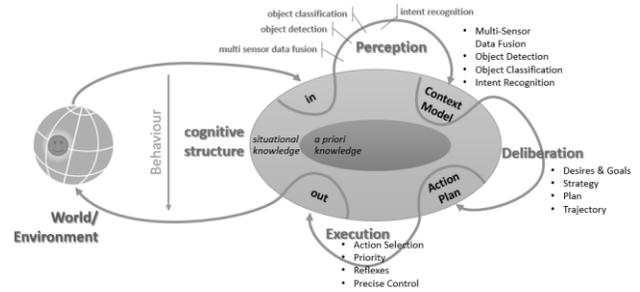

Figure 1: Cognitive Process

The body of the cognitive system is structured into the a-priori-knowledge (inner and darker oval in Figure 1) and situational knowledge (outer oval). The a priori knowledge is generated by the knowledge transformators using their a-priori knowledge during runtime. This can be understood like an instantiation process of the concepts within the a-priori knowledge. To do this, each transformator "reads" primarily in its input area (but may use the whole knowledge) and writes into the output area. So, the Perception reads primarily from "in" and writes to "Context Model" which describes the analyzed situation the system is in. This might include a representation of objects in the environment but also might comprise of abstract objects like distances or threads. The Deliberation takes the Context Model and determines what to do and how. A (structured) action plan is the result of the Deliberation. Last but not least the Execution takes the Action Plan as primary input, selects current actions, takes priorities into account and writes commands to the out area. The commands in the out area are taken and processed by the output to drive actuators and to manipulate the environment. This model of a cognitive system represents working model and functional abstraction for structured approach (trustworthiness reference lifecycle) discussed in this work.

An A/C-system might include a subset of the following characteristics:

- recognizes its environment (or parts of it) through "sensors",
- knows about the intentions of elements in its environments (e.g. implements intent recognition),
- knows about higher level goals (might even incorporate ethical point of views),
- takes (non-trivial) decisions based on reasoning,
- influences its environment via actuators (distinguish from actor in the sense of performer),
- interacts and cooperates with the elements of its environment,

- influences elements in its environment to better meet its own goals (e.g. mechanism design),
- shows a certain behavior based on skills, and
- learns even new behavior during runtime.

Examples for A/C-systems are Advanced Driver Assistance Systems, Automated vehicles or Autonomous Robots.

**Trustworthiness and its Analysis**

The term trustworthiness has no generally accepted definition. This work considers trustworthiness as a more generic concept that combines a user defined and potentially project specific set of aspects. These aspects include but are not limited to (functional) safety, security, privacy, usability, ethical and legal compliance, reliability, availability, maintainability, and (intended) functionality (see Figure 2).

These non-functional properties, forming trustworthiness concept, are brought into the system by applying certain methods and by the way in which the original functional requirements are implemented. Therefore these characteristics (non-functional properties) are called "emerging", i.e. not directly implementable. They need to be built into the product during design time. Furthermore, these characteristics need to be proven on the basis of process documentation, the use and implementation of suitable methods and measures and finally, by the capability of the designers. In order to address these issues, a structured process needs to be followed throughout the whole design cycle of an A/C-system (see Section – Reference Lifecycle), including the components that contain NNs or other AI algorithms.

Another challenge is to balance between all potentially conflicting aspects. For example, safety and security might support or exclude each other. And traditionally, aspects of security and usability do conflict. These need to be resolved and balanced decisions need to be taken.

Implementing a trustworthy system of interest follows the well-known approach along the reference lifecycle of the following steps:
- analyze (trustworthiness) hazards and assess (trustworthiness) risks
- define a (trustworthiness) concept consisting of (trustworthiness) mitigation measures
- implement (trustworthiness) concept

In every step of a lifecycle, trustworthiness is considered with the same set of aspects (and scopes). During the trustworthiness analysis, special care has to be taken by combining its aspects (safety, security, etc.), especially when hazard attributes like integrity and assurance are combined or when combining aleatoric and epistemic uncertainties. During the implementation phases this is less relevant. There the specific performance of the implemented element is the relevant issue. It relates to integrity/assurance/uncertainties trough traceability, but these are no direct input during e.g. the implementation of a software unit. To reflect this observation the reference lifecycle takes into account "Trustworthiness Performance Level" (TPL) as a one-dimensional performance attribute of trustworthiness requirements.

## Reference Lifecycle

This section presents an approach towards trustworthy A/C-systems. The backbone of this approach is a system lifecycle discussed in this section which is the core of further contributions such as addition of the Solution Level, and concerns related to AI in a form of AI Design and AI Blueprint concepts.

### Overview

An assurance case is a convincing and structured argumentation based on evidences that the A/C-system is sufficiently trustworthy. Trustworthiness with every aspect like safety, security etc. (see Figure 2) is an emergent property of a system of interest. It emerges from all activities during the engineering phase. So the structured argument that trustworthiness is met needs to be based on a structured approach of the engineering phase, which motivates this work. One of the main problems that prevents A/C-systems from being qualifyable, is an unstructured and ad-hoc way of developing them, especially components that contain AI technology. These have negative impact on the ability to compile a structured argument of trustworthiness based on evidences. While for classical software and hardware, process models that ensure certain level of rigor have been developed , these are not directly applicable to AI specific systems. This motivates a definition of a structured lifecycle which accounts also for AI concerns.

The reference lifecycle discussed in this work covers the Design part of the overall product lifecycle (see Figure 3).

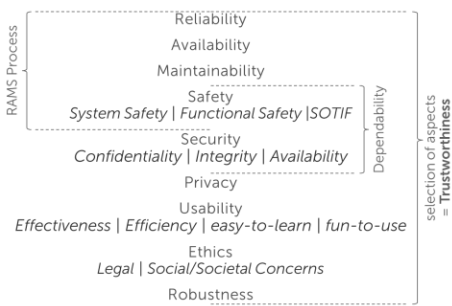

Figure 2: Aspects forming the metaterm "Trustworthiness"

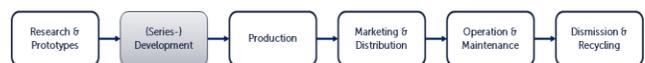

Figure 3: Product Lifecycle

Figure 4 presents the reference lifecycle. The name, reference lifecycle, signifies its main purpose, i.e. it can be used as a reference for a standard that supports assessment approach for trustworthy systems. It is inspired by and resembles to certain degree the structure of the safety lifecycle in IEC 61508 and similarly in ISO 26262. However, this approach is not restricted to it and can be adapted to other standards like IEC 61508, ISO/IEC 15504 (Software Process Improvement and Capability, SPICE) [ISO/IEC (2008b)], and ISO/IEC 12207 (Systems and Software Engineering – Software lifecycle processes) [ISO/IEC (2008a)]. Apart from the process structure, this framework absorbs also the ISO 26262-like safety argumentation based on the integrity principle

The reference lifecycle is a phase model that arranges concepts (e.g. Initiation) into logical dependency sequence. The reference lifecycle is not a process model. The phases are not to be understood as a waterfall model. The reference lifecycle defines the logical flow of activities but is open to any actual process model (e.g. waterfall, V model, spiral model). This flow of activities grouped in phases of the reference lifecycle lead to the design and implementation of the solution accompanied by its trustworthiness assurance case.

Starting from the top, first is Initiation phase, which objects in finding the solution that shall be developed. At that stage, interfaces, environment and usage of the solution and last but not least, requirements concerning trustworthiness shall be understood. Also, synchronization with organization and process framework shall be executed, and finally, competent team to work on different aspects of the product shall be setup. Concerning principia of this phase, it can be related to the Item definition and Initiation of the safety lifecycle phases, as defined in ISO 26262.

Next, development at Solution Level expresses one of the contributions of the overall lifecycle, hence it is described in more details in a separate section.

Development at *System Level* is also characteristic to other standards, listed before, with an exception that the input to that phase in proposed reference lifecycle comes from the *Solution Level*, which is not present in most other standards. From that standpoint, system level as defined here has additional, unique characteristics, which are discussed more broadly in section – System Level.

Activities of the Technology Level are focused on contributing to the solution, based on a certain type of technology. Here, refinement of trustworthiness concept into specific technology that will implement it, takes place. For hardware and software this lifecycle refers to the corresponding, well defined and described activities, specified in existing standards. For instance, ISO 26262 part 5 thoroughly describes development at hardware level, and correspondingly, part 6 does the same for software. Contribution of the trustworthiness lifecycle is that it puts into consideration AI, and hence introduces two concepts at this phase, i.e.

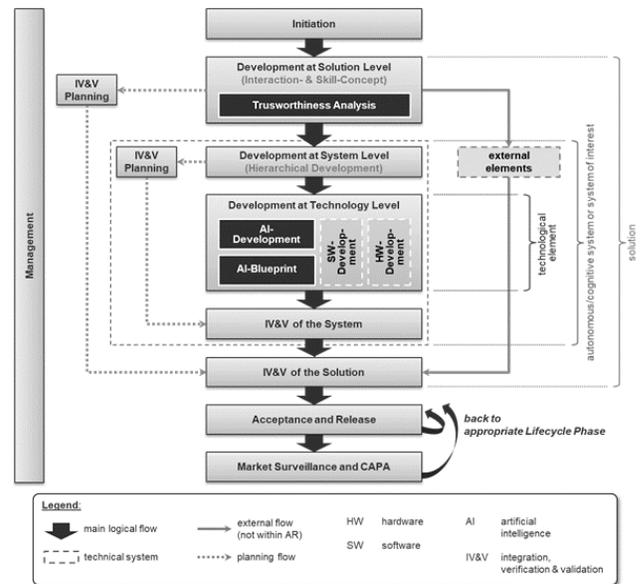

Figure 4: Reference Lifecycle

AI Design and AI Blueprint. These are discussed in more details, correspondingly in section AI Development and AI Blueprint.

The IV&V of the System refers to the concerns of design verification and validation especially in terms of its compliance and completeness with regards to the technical trustworthiness concept. This implies usage of such methods as system design inspection, walkthrough, simulation, and from the trustworthiness perspective, this is trustworthiness analysis. The similar purpose as of the previous phase, guides activities identified for the IV&V of the Solution. Namely, solution is inspected to verify and validate its compliance and completeness in regards to functional and trustworthiness requirements.

Next phase, i.e. Acceptance and Release covers several objectives. This is preparation of the release documentation which specifies, inter alia, criteria for the release for production. Next, this is compilation of a trustworthiness assurance case, i.e. how, over the reference lifecycle, trustworthiness objectives were reached. This requires to deeply and thoroughly assess the trustworthiness of the solution concept. Ultimate objective is to release the solution for production.

Last step, with tight correspondences to existing standards, concerns market surveillance and CAPA (Corrective Action Preventive Action). These are all activities which focus on the monitoring of a product in its operational environment and necessary reactions to possible malfunctioning and dissatisfaction of end users resulting from it.

### Solution Level

An A/C-system is more than an algorithm that is categorized as artificial intelligence and that is implemented in one or more elements of the A/C-system. To understand the full complexity it is necessary to take the user of the A/C-system into account as well as all relevant interfaces and elements in the environment. Actually this is the level of abstraction that is called the solution level. The solution level delineates the whole setting that can be perceived as an architecture in which the A/C-system is one element.

The intent of the Solution Level is to generate a solution concept on the basis of all customer requirements which might - in the first place - include conflicts. Resolving conflicts and including the use and environment of the product results in a consistent solution concept. Furthermore the solution level is the relevant origin of all hazards, because in most cases hazards do not arise from a system of interest itself but from setting the system into an environment.

The definition of the solution concept refers to a black box model and a white box model:

- The **black box model** focuses on the interfaces, the behavior (including interaction and cooperation of the A/C-system with other elements in the solution) and further requirements of the A/C-system. One of the crucial elements of this black box model is to define the task share between the A/C-system, the user and other elements. A typical black box model is the sociotechnical work system.
- The **white box model** takes a look into the A/C-system as an element of the solution level architecture. At this phase of the development the description is kept at a very abstract level and remains functional in most cases. The purpose of the white box model is to better understand the behavioral elements of the A/C-system. For this purpose the behavior defined using the black box model is detailed to skills of the A/C-system describing the mechanisms of the recognize-act-cycle as the closed loop between environment and A/C-system. An example for a white box model is the generic sense-plan-act model.

At this level the development team shall review the customer requirements, and all other relevant material to understand the scope and goals of the solution. Then, the team shall describe the overall solution using the defined notation. This description includes the goals (e.g. in terms of use cases) of the development, the environment and its relevant elements including the user, the interfaces of the system to all relevant elements in the environment, and the boundaries of the system.

Next, the same team needs to define the black box model to describe the observable behavior of the system – item. This description should cover all relevant aspects of the interaction and cooperation within the solution. It also needs to define the black box behavior (input-output-mapping that can be observed), and allocate behavioral requirements to other elements in the solution. This black box model shall be then used to describe the interaction and cooperation concept. This includes description of overall tasks of the solution, task share between item and other elements of the solution (user or machine, etc.), interaction and cooperation with other elements, definition of the black box behavior of the item (input-output-mapping that can be observed), and allocation of behavioral requirements to other elements in the solution.

Following is the specification of the solution behavior, described as functional chains and including necessary skills in the machine. This is the mentioned white box model. Such a description shall include functional architecture as a partition of the behavior (recognize-act-cycle), cognitive theory on how to generate behavior (= mapping between input and output), and cognitive architecture description.

The development team shall then use the white box model to describe the internal processes of the system. This description shall include functional architecture of the system, definition of elements according to the white box model (e.g. behavior & skills of the system), definition of interfaces between these elements, and collaboration of these elements via the given interfaces to generate behavior.

Ultimate outcome of the activities performed at the solution level should include solution definition, cooperation concept, functional architecture of the system, and acceptance criteria for cooperation and behavior. The last is required aspect of trustworthiness concept.

**System Level**

The development at system level is the link between the definition of the solution (solution level) and the implementation according to a certain technology (technology level). In the first place there are no trustworthiness specific characteristics or activities. These systems engineering activities can be organized according to typical systems engineering standards (e.g. ISO/IEC 15288). The people in charge of the development at system level should care for a good design and the application of organizational, proven and state-of-the-art processes, supporting processes, architectures, methods and measures.

Taking a detailed look at the system level, the awareness of trustworthiness induces at least two aspects that are relevant to achieving trustworthiness in the resulting A/C-system and solution:

- repetitive / iterative application of that part of the standard to cover the complexity of system-of-systems as well as complex system architectures
- traceability of trustworthiness attributes throughout all levels of the design hierarchy, including methods like allocation, decomposition and segregation
- compilation of the trustworthy assurance case, and
- design patterns that support verification and AI properties

The hierarchical design will be very specific to the A/C-system, its functionality and its domain. For example in the domain of embedded systems the functionality along with hardware and software are closely related even during the development. This is the nature of embedded systems. For them one would expect optional system-of-system level in the higher levels of the hierarchical design, one or more levels where the sub systems are handled and finally a level where electronic control units (ECU) are defined and detailed into the technology level using hardware, software and (some of the ECUs) AI.

**AI Development**

Activities of the Technology Level are focused on contributing to the solution, based on a certain type of technology. The design of components based on SW or HW is considered separately within existing standards, due to the different concerns that these technologies raise. This work argues that unique characteristics of AI technology predestine it for having separate place within the overall lifecycle. This implies establishment of AI Development phase.

Design and implementation of AI-based solution resembles to some extent design and implementation as done with standards SW approach. Simply, most of the ML (machine learning) algorithms is implemented in SW and therefore AI gained perception similar to SW. Nevertheless, AI could definitely be perceived as additional abstraction on top of the SW. It could be compared to model driven development (which ultimately also relies on SW), but it comes along with the new philosophy for design and implementation, different kind of tooling, and hence different requirement in terms of knowledge, useful to work with these technologies. For MDD this means development of models out of which code is being generated and then compiled. New abstraction definitely brings new challenges but also opportunities. The latter in terms of AI this is mainly possibility to implement features operating in open context (e.g. autonomous driving). This however imposes challenges, which are not characteristic to pure SW development. For example, proper AI algorithm needs to be selected. This step could be compared to justified selection of programming language, which shall serve best the implementation of intended functionality, respecting for instance non-functional concerns such as execution time. Next, if neural networks are used, their design (e.g. choice of the number of layers) could be referred to SW architecture design. However, both solve different problems and definitely require different set of skills. Another, significant difference refers to ultimate implementation. Namely, in SW this boils down to the coding of atomic SW units specified as part of the SW architecture. For NN, this is learning process. How SW units are implemented depends on the requirements which are attached to them. On the other hand, there is no explicit requirements specification for NN. Requirements are implicitly embodied by the collected data, used for the learning process, and this data ultimately impacts the NN implementation. This is definitely a relevant difference between SW and AI, one that has also huge implications on how trustworthiness is considered. Having standard requirements which then could be attached to SW units, enables proper traceability. This is not possible for NN. In conclusion, AI reveals new challenges when compared to SW, both from the design but also trustworthiness perspective. This drives the idea of treating its design as a separate concern within the overall framework.

There are several objectives of AI Development. The first and most important is the selection of AI technology which is believed to provide the best solution for the required functionality. This decision has a relevant impact on the following activities. One of them is a choice of an appropriate AI Blueprint (see following section) and then its adaptation (tuning) and application so that it can be effectively used with the selected AI technology. Having these preconditions, ultimate objective, i.e. delivery of the AI element together with all necessary documentation and qualifications to the system integration and verification can be fulfilled.

Inputs to AI Development phase these are functional and non-functional requirements allocated to the element, and trustworthiness attributes. The first drive the design of a solution. The second has substantial impact on the reasoning about the trustworthiness qualification of the designed AI component.

**AI Blueprint**

The development or training of AI components does not fit into existing process models (e.g. like for classical software) due to the specific nature of the AI methodologies. Even inside the field of AI, different methodologies and solution concepts can have very specific requirements towards the underlying process model. For example, a deep convolutional neural network for supervised learning has almost nothing in common with STRIPS planning, a purely symbolic automated planning approach. This urges for the new approach in which specific characteristics of certain AI technology are targeted by design process. In consequence, the framework introduces concept of AI blueprint and the need of defining specific blueprint, depending on the type of AI technology being used. This means that standard should be flexible enough to enable incorporation of blueprints which use specific methods or narrow set of methods.

The blueprint can be interpreted as a kind of template process that can be applied to the relevant kind of AI methodology. It is characterized by Input and Output Interfaces, Structure (i.e. design phases) and Qualifications. The execution of AI blueprint provides an AI element characterized by a predefined quality level, including guarantee to meet defined dependability requirements.

In order for the blueprints itself to be incorporable within the overall design lifecycle, there exist two different forms of requirements imposed on them. The first one is the technical requirement, that all blueprints possess required interfaces, in order to plug them into the development lifecycle.

This means that they process the properties and artifacts that are handed down to them, as well as that they deliver the required results back to the higher levels.

As inputs of the AI development the requirements and trustworthiness requirements with additional attribute like TPL are provided by the system level, and as output of the AI development the system level expects the AI element along with the confidence level $\lambda_{AI}$, which conceptually is similar to $\lambda$ for HW (from ISO 26262), but requires different approach to calculate it.

Figure 5 presents an example of an AI Blueprint dedicated to develop AI Element, using NNs and supervised learning. In this case, similarly as it is advocated in ISO 26262 for SW or HW design, this AI Blueprint is based on the V-model. The last of course is not the required property of AI Blueprint. It adheres to the above requirements, i.e. it accepts two input elements, i.e. trustworthiness requirements and TPLs. Similarly, an output of this AI Blueprint delivers AI element together with confidence level $\lambda_{AI}$.

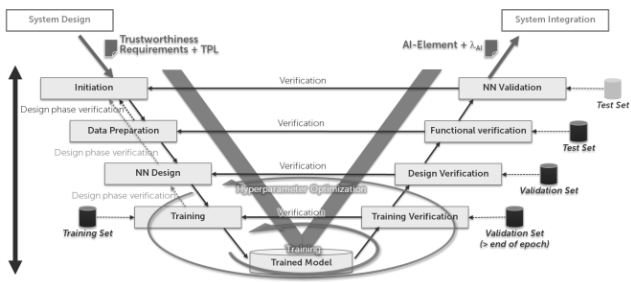

Figure 5: AI Blueprint for Supervised Learning of NNs

## Summary and Future Work

This work described trustworthiness reference lifecycle. The main contribution of this work in regards to existing approaches or standards lies in the overall structure of the trustworthiness lifecycle. These are especially:

- addition of one level above the engineering of the system of interest, i.e. solution level
- enablement of approaches based on integrity and assurance
- introduction of "AI" as a 3$^{rd}$ kind besides software and hardware, resulting in the concepts of AI Development and AI Blueprint.

There are several parts of the overall framework which still require deeper consideration and explanation. These will be considered in the following series of papers. This is more exhaustive specification of the AI Blueprint concept, where the most important is introduction and deeper discussion of examples of blueprints that could be used. In parallel, deep discussion over the definition of the confidence parameter will be presented. The last will allow to reason about the trustworthiness qualifications of the AI design. This is so called $\lambda_{AI}$ concept, briefly introduced in this paper.